\documentclass[twocolumn]{article}

\usepackage{pstricks}
\usepackage{graphicx,color}
\usepackage[left=2cm,top=2cm,right=2cm,bottom=2cm,nohead]{geometry}
\usepackage{graphics,latexsym,amsmath,amssymb,xspace,enumerate,paralist}
\usepackage{wrapfig,subfigure}
\usepackage[noend]{algorithmic}
\usepackage{algorithm}

\newcommand{\scrA}{\ensuremath{\mathcal{A}}}

\newcommand{\scrD}{\ensuremath{\mathcal{D}}}
\newcommand{\scrE}{\ensuremath{\mathcal{E}}}
\newcommand{\scrF}{\ensuremath{\mathcal{F}}}
\newcommand{\scrS}{\ensuremath{\mathcal{S}}}
\newcommand{\scrU}{\ensuremath{\mathcal{U}}}
\newcommand{\scrX}{\ensuremath{\mathcal{X}}}
\newcommand{\scrY}{\ensuremath{\mathcal{Y}}}
\newcommand{\scrZ}{\ensuremath{\mathcal{Z}}}
\newcommand{\rA}{\ensuremath{\rightarrow}}
\newcommand{\rrA}{\ensuremath{\longrightarrow}}

\newcommand{\XB}{\ensuremath{\mathbf{X}}}
\newcommand{\zeroB}{\ensuremath{\mathbf{0}}}
\newcommand{\sm}{\mbox{\textendash}}

\newcounter{pseudocode}

\newtheorem{theorem}{Theorem}[section]

\newtheorem{definition}[theorem]{Definition}

\newtheorem{example}{Example}

\newcommand {\comment}[1]{}

\newlength\colwidth

\newcount\hours
\newcount\minutes
\def\SetTime{\hours=\time
        \global\divide\hours by 60
        \minutes=\hours
        \multiply\minutes by 60
        \advance\minutes by-\time
        \global\multiply\minutes by-1 
        \ifnum\hours<12 \def\ampm{am} 
        \ifnum\hours<1 \advance\hours by+12 \fi
        \else
        \def\ampm{pm} \advance\hours by-12 \fi }
\SetTime

\def\now{\number\hours:\ifnum\minutes<10 0\fi\number\minutes\ampm}

\makeatletter
\def\cramped                                    
  {\parskip\@outerparskip\@topsep\parskip       
  \@topsepadd2pt\itemsep0pt                     %
}
\makeatother

\begin{document}
\title{Inferring Dynamic Bayesian Networks\\
using Frequent Episode Mining}

\author{Debprakash Patnaik$^\dagger$, Srivatsan Laxman$^*$, and Naren Ramakrishnan$^\dagger$\\
$^\dagger$Department of Computer Science, Virginia Tech, VA 24061, USA\\
$^*$Microsoft Research, Sadashivanagar, Bangalore 560080, India}

\date{} \maketitle
\begin{abstract}
\noindent
{\bf Motivation:} Several different threads of research have been proposed for modeling and mining temporal data. On the one hand, approaches such as dynamic Bayesian networks (DBNs) provide a formal probabilistic basis to model relationships between time-indexed random variables but these models are intractable to learn in the general case. On the other, algorithms such as frequent episode mining are scalable to large datasets but do not exhibit the rigorous probabilistic interpretations that are the mainstay of the graphical models literature.

\noindent
{\bf Results:} We present a unification of these two seemingly diverse threads of research, 
by  demonstrating how dynamic (discrete) Bayesian networks can be inferred from the 
results of frequent episode  
mining. This helps bridge the modeling emphasis of the former with the counting emphasis of the latter. 
First, we show how, under reasonable assumptions on data characteristics and on influences of
random variables, the optimal DBN structure can be computed using a greedy, local, algorithm.
Next, we connect the optimality of the DBN structure
with the notion of fixed-delay episodes and their counts of distinct occurrences. Finally, 
to demonstrate the practical feasibility of our approach,
we focus on a specific (but broadly applicable) class of networks, called excitatory networks,
and show how the search for the optimal DBN structure can be conducted using 
just information from frequent episodes. Application on datasets gathered from mathematical models of spiking
neurons as well as real neuroscience datasets are presented.

\noindent
{\bf Availability:} Algorithmic implementations, simulator codebases, and datasets are available from our website at http://\hskip0ex neural-code.cs.vt.edu/dbn.

\end{abstract}

%
%
\vspace{1mm}



\section{Introduction}
\label{intro}
Probabilistic modeling of temporal data is a thriving area of research. The development
of dynamic Bayesian networks as a subsuming formulation to HMMs, Kalman filters, and other
such dynamical models has promoted a succession of research aimed at capturing
probabilistic dynamical behavior in complex systems. DBNs bring to modeling
temporal data the key advantage that traditional Bayesian networks brought to modeling
static data, i.e., the ability to use graph theory to capture probabilistic notions of
independence and conditional independence. They are now widely used in bioinformatics,
neuroscience, and linguistics applications.

A contrasting line of research in modeling and mining temporal data is the counting based
literature, exemplified in the KDD community by works such as~\cite{MTV97,vatsan1}. Similar
to frequent itemsets, these papers define the notion of frequent episodes as objects of interest.
Identifying frequency measures that support efficient counting procedures (just as support
does for itemsets) has been shown to
be crucial here.

It is natural to question whether these two threads, with divergent origins, can be
related to one another. 
Many researchers have explored precisely this question.
The classic paper by Pavlov, Mannila, and Smyth~\cite{pavlov} used frequent
itemsets to place constraints on the joint distribution of item random variables
and thus aid in inference and query approximation. 
Chao and Parthasarathy~\cite{chaop} view probabilistic models as summarized
representations of databases and demonstrate how to construct MRF (Markov random
field) models from frequent itemsets.
Closer to the topic of this paper, the work by Laxman et al.~\cite{vatsan1}
linked frequent episodes to a generative HMM-like model of the underlying data.

Similar in scope to the above works, we present a 
unification of the
goals of dynamic Bayesian network inference with that of frequent episode mining.
Our motivations are not merely to establish theoretical results but also to
inform the computational complexity of algorithms and spur faster
algorithms targeted for specific domains.
The key contributions are:
\begin{enumerate}
\item We show how, under reasonable assumptions on data characteristics and on influences of
random variables, the optimal DBN structure can be established using a greedy, local, approach,
and how this structure can be computed using the notion of fixed-delay episodes and their
counts of distinct occurrences. 
\item We present a 
specific (but broadly applicable) class of networks, called excitatory networks,
and show how the search for the optimal DBN structure can be conducted using just information
from frequent episodes. 
\item We demonstrate a powerful
application of our methods on datasets gathered from mathematical models of spiking
neurons as well as real neuroscience datasets.
\end{enumerate}

%
%
%
%
%
%
%
%
%
%
%
%
%
%

\section{Bayesian Networks: Static and Dynamic}
Formal mathematical notions are presented in the next section, but here
we wish to provide some background context to past research in Bayesian networks (BNs).
As is well known, BNs use directed acyclic graphs to encode
probabilistic notions of conditional independence, such as that a node is
conditionally independent of its non-descendants given its parents (for more details,
see~\cite{jordan-book}). The earliest
known work for learning BNs is the Chow-Liu algorithm~\cite{cl-trees}.
It showed that, if we restricted the structure of the BN to be a tree, 
then the optimal BN can be computed using a minimum spanning tree algorithm.
It also established the tractability of BN inference for this class of graphs.

More recent work, by Williamson~\cite{williamson00}, generalizes the Chow-Liu
algorithm to show how (discrete) distributions can be generally approximated using
the same ingredients that are used by the Chow-Liu approach, namely mutual information
quantities between random variables. Meila~\cite{meila} presents an accelerated algorithm 
that is targeted toward sparse datasets of high dimensionality. The approximation 
thread for general BN inference is perhaps best exemplified by Friedman's sparse candidate
algorithm~\cite{FMR98} that presents various approaches to greedily learn (suboptimal)
BNs.

Dynamic Bayesian networks are a relatively newer development and best examples of them
can be found in specific state space and dynamic modeling contexts, such as HMMs.
In contrast to their static counterparts, exact
and efficient inference for general classes of DBNs has not been well studied.

\section{Optimal DBN structure}
\label{sec:optimal-dbn-structure}

Consider a finite alphabet, $\scrE = \{A_1,\ldots,A_M\}$, of event-types (or symbols). Let
$s=\langle (E_1, t_1), (E_2,t_2), \ldots, (E_n,t_n) \rangle$ denote a data stream of $n$ events over $\scrE$.
Each $E_i, i=1,\ldots,n$, is a symbol from $\scrE$. Each $t_i, i=1,\ldots,n$, takes values from the set of positive integers.
The events in $s$ are ordered according to their times of occurrence, 
$t_{i+1}\geq t_i, i=1,\ldots,(n-1)$. The time of occurrence of the last event in
$s$, is denoted by $t_n=T$. We model the data stream, $s$, as a realization of a
discrete-time stochastic process $\XB(t), t = 1,\ldots,T$; $\XB(t)=[X_1(t) X_2(t)\cdots X_M(t)]'$,
where $X_j(t)$ is an indicator variable for the occurrence of event type, $A_j\in\scrE$, at time
$t$. Thus, for $j=1,\ldots,M$ and $t=1,\ldots,T$, we will have $X_j(t)=1$ if $(A_j,t)\in s$, and $X_j(t)=0$ otherwise.  
Each $X_j(t)$ is referred to as the {\em event-indicator random variable} for event-type, $A_j$, at time
$t$.
\begin{example}
\label{ex:data-example}
The following is an example event sequence of $n=7$ events over an alphabet, $\scrE=\{A,B,C,\ldots,Z\}$, of
$M=26$ event-types:
\begin{equation}
\langle (A,2), (B,3), (D,3), (B,5), (C,9), (A,10), (D,12) \rangle
\label{eq:ex-seq}
\end{equation}
The maximum time tick is given by $T=12$. Each $\XB(t),\ t = 1,\ldots,12$, is a vector of $M=26$
indicator random variables. Since there are no events  at time $t=0$ in the example sequence (\ref{eq:ex-seq}), we have
$\XB(1)=\zeroB$. At time $t=2$, we will have $\XB(2)=[1 0 0 0 \cdots 0]'$. Similarly, $\XB(3)=[0 1 0 1 \cdots
0]'$, and so on.
\end{example}

A Dynamic Bayesian Network \cite{murphy02} is essentially a DAG with nodes representing random variables and arcs representing
conditional dependency relationships. In this paper, we model the random process $\XB(t)$ (or equivalently, the
event stream $s$), as the output of a Dynamic Bayesian Network. Each event-indicator, $X_j(t)$, $t=1,\ldots,T$ and $j=1,\ldots M$,
corresponds to a node in the network, and is assigned a set of parents, which is denoted as $\pi(X_j(t))$ 
(or $\pi_j(t)$). A parent-child relationship is represented by an arc (from parent to child) in the DAG.  In 
a Bayesian Network, nodes are conditionally  independent of their non-descendants given their
parents. The joint probability distribution of the random process, $\XB(t)$, under the DBN model,
can be factorized as a product of conditional probabilities, $P[X_j(t)\:|\:\pi_j(t)]$, for various $j,t$.

In general, given a node, $X_j(t)$, any other $X_k(\tau)$ can belong to its parent
set, $\pi_j(t)$. 
However, since each node has a time-stamp, 
it is reasonable to assume that a random variable, $X_k(\tau)$, can only influence {\em future}
random variables (i.e.~those random variables associated with later time indices). Also, we can expect
the influence of  $X_k(\tau)$ to diminish with time, and so we assume that
$X_k(\tau)$ can be a parent of $X_j(t)$ {\em only if} time $t$ is within $W$ time-ticks of time
$\tau$ (Typically, $W$ will be small, like say, 5 to 10 time units). 
All of this constitutes our first constraint, {\bf A1}, on the DBN structure.
\begin{description}
\item[A1]: For user-defined parameter, $W > 0$, 
the set, $\pi_j(t)$, of parents for the node, $X_j(t)$, is a subset of event-indicators out of the
$W$-length history at time-tick, $t$, 
i.e.~$\pi_j(t) \subset \{X_k(\tau)\::\:1\leq k \leq M,\ (t-W)\leq \tau <t\}$.
\end{description}
The DBN essentially models the time-evolution of the event-indicator random variables associated
with the $M$ event-types in the alphabet, $\scrE$. By learning the DBN structure, we
expect to unearth relationships like ``event $B$ is more likely to occur at time $t$, if $A$ occurred 3
time-ticks before $t$ and if $C$ occurred 5 time-ticks before $t$.'' In view of this, it is reasonable to assume 
that the parent-child relationships depend on relative (rather than absolute) time-stamps of random variables
in the network. We refer to this as {\em translation invariance} of the DBN structure, and is
specified below as the second constraint, {\bf A2}, on the DBN structure.
\begin{description}
\item[A2]: If $\pi_j(t) = \{X_{j_1}(t_1),\ldots,X_{j_\ell}(t_\ell)\}$ is an $\ell$-size parent
set of $X_j(t)$ for some $t>W$, then for any other $X_j(t'),\ t'>W$, its parent set, $\pi_j(t')$, is simply a
time-shifted version of $\pi_j(t)$, and is given by $\pi_j(t') =$ $\{X_{j_1}(t_1+\delta),\ldots,$ 
$X_{j_\ell}(t_\ell+\delta)\}$, where $\delta = (t' - t)$.
\end{description}
The data stream, $s$, is a long stream of events which we will regard as a realization of
the stochastic process, $\XB(t),\ t=1,\ldots,T$. While {\bf A2} is a sort of structural
stationarity constraint on the DBN, in order to estimate marginals of the joint distribution from the data
stream, we will also require that the distribution does not change when shifted in time.
The stationarity assumption is stated in {\bf A3} below.
\begin{description}
\item[A3]: For all $j$, $\delta$, given any set of $\ell$ event-indicators, say, $\{X_{j_1}(t_1),
\ldots,X_{j_\ell}(t_\ell)\}$, the stationarity assumption requires that,
$P[X_{j_1}(t_1), \ldots, X_{j_\ell}(t_\ell)] = P[X_{j_1}(t_1+\delta),$ $\ldots, X_{j_\ell}(t_\ell+\delta)]$.

\end{description}
The joint probability distribution, $Q[\cdot]$, under the Dynamic Bayesian Network model can
be written as:
\begin{align}
Q[\XB(1),\ldots,\XB(T)] &= P[\XB(1),\ldots,\XB(W)] \nonumber \\
&\times \prod_{t=W+1}^T \prod_{j=1}^M P[X_j(t)\:|\:\pi_j(t))]
\label{eq:dbn}
\end{align}
Learning the structure of the network involves learning the map, $\pi_j(t)$, for each $X_j(t)$,
$j=1,\ldots,M$ and $t=(W+1),\ldots,T$. In this paper, we derive an {\em optimal} structure for a 
Dynamic Bayesian Network, given an event stream, $s$, under assumptions {\bf A1}, {\bf A2} 
and {\bf A3}. Our approach follows the lines of \cite{cl-trees,williamson00} where structure learning is
posed as a problem of approximating the discrete probability distribution, $P[\cdot]$, by the best
possible distribution from a chosen model class (which, in our case, is the class of Dynamic Bayesian
Networks constrained by {\bf A1} and {\bf A2}). The Kullback-Leibler
divergence between the underlying joint distribution, $P[\cdot]$, of the stochastic process, and the
joint distribution, $Q[\cdot]$, under the DBN model is given by
\begin{align}
D_{KL}(P || Q) = \sum_\scrA & \bigg( P[\XB(1),\ldots,\XB(T)] \nonumber \\
&\times \log \frac{P[\XB(1),\ldots,\XB(T)]}{Q[\XB(1),\ldots,\XB(T)]} \bigg)
\end{align}
where $\scrA$ represents the set of all possible assignments for the $T$ $M$-length random vectors,
$\{\XB(1),\ldots,\XB(T)\}$. Denoting the entropy of $P[\XB(1),\ldots,\XB(T)]$ by $H(P)$, the entropy of
the marginal, $P[\XB(1),\ldots,\XB(W)]$, by $H(P_W)$, and substituting for $Q[\cdot]$ from Eq.~(\ref{eq:dbn}), we get
\begin{align}
D_{KL}(P||Q) = -& H(P) - H(P_W) - \sum_\scrA \bigg( P[\XB(1),\ldots,\XB(T)] \nonumber\\
&\times \sum_{j=1}^M \sum_{t=W+1}^T \log P[X_j(t)\:|\:\pi_j(t)]\bigg) \label{eq:DKL1}
\end{align}
We now expand the conditional probabilities in Eq.~(\ref{eq:DKL1}) using Bayes rule, switch the order of summation
and marginalize $P[\cdot]$ for each $j,t$. Denoting, for each $j,t$, the entropy of the marginal $P[X_j(t)]$
by $H(P_{j,t})$, the expression for KL divergence now becomes:
\begin{align}
D_{KL}(P||Q)  = - H(P) &- H(P_W) - \sum_{j=1}^M \sum_{t=W+1}^T H(P_{j,t})\nonumber\\
&- \sum_{j=1}^M \sum_{t=W+1}^T I[X_j(t)\:;\:\pi_j(t)]
\label{eq:DKL2}
\end{align}
$I[X_j(t)\:;\:\pi_j(t)]$ denotes the {\em mutual information} between $X_j(t)$ and its parents,
$\pi_j(t)$, and is given by
\begin{align}
I[X_j(t)\:;\:\pi_j(t)] = &\sum_{\scrA_{j,t}} \bigg( P[X_j(t),\pi_j(t)] \nonumber \\
& \times \log \frac{P[X_j(t),\pi_j(t)]}{P[X_j(t)]\ P[\pi_j(t)]} \bigg)
\label{eq:mutual-information}
\end{align}
where $\scrA_{j,t}$ represents the set of all possible assignments for the random variables,
$\{X_j(t), \pi_j(t)\}$. Under the translation invariance constraint, {\bf A2}, and the stationarity
assumption, {\bf A3}, we have $I[X_j(t)\:;\:\pi_j(t)] = I[X_j(t')\:;\:\pi_j(t')]$ for
all $t>W$, $t'>W$. This gives us the following final expression for $D_{KL}(P||Q)$:
\begin{align}
D_{KL}(P||Q)  = &- H(P) - H(P_W) - \sum_{j=1}^M \sum_{t=W+1}^T H(P_{j,t})\nonumber\\
&- (T-W) \sum_{j=1}^M I[X_j(t)\:;\:\pi_j(t)]
\label{eq:DKL3}
\end{align}
where $t$ is any time-tick satisfying $(W < t \leq T)$. We note that in Eq.~(\ref{eq:DKL3}), the entropies,
$H(P)$, $H(P_W)$ and $H(P_{j,t})$ are independent of the DBN structure (i.e.~they do not depend
on the $\pi_j(t)$ maps). Since $(T-W)>0$ and since $I[X_j(t)\:;\:\pi_j(t)]\geq 0$ always, the KL
divergence, $D_{KL}(P||Q)$,  is minimized when the sum of $M$ mutual information terms
in Eq.~(\ref{eq:DKL3}) is maximized.  Further, from {\bf A1} we know that all parent nodes of
$X_j(t)$ have time-stamps strictly less than $t$, and hence, no choice of $\pi_j(t),\
j=1,\ldots,M$ can result in a cycle in the network (in which case, the structure will not be a DAG,
and in-turn, it will not represent a valid DBN).  This ensures that, under the restriction of {\bf A1}, 
the {\em optimal DBN structure} (namely, one that corresponds to a $Q[\cdot]$ that minimizes KL divergence with respect to the true joint
probability distribution, $P[\cdot]$, for the data) can be obtained by {\em independently} picking the highest mutual
information parents, $\pi_j(t)$, for each $X_j(t)$ for $j=1,\ldots,M$ (and, because of {\bf A2} and
{\bf A3}, we need to carry-out this parents' selection step only for the $M$ nodes in any one time slice, $t$, that satisfies $(W<t\leq T)$).


\section{Fixed-delay episodes}
\label{sec:episodes}

Frequent episode discovery is a popular framework in temporal data mining \cite{LSU07a,PSU08,BCSZ06}.
The data in this framework is a single long stream of events over a finite alphabet, as defined at
the beginning of Sec.~\ref{sec:optimal-dbn-structure} (cf.~{\em Example~\ref{ex:data-example}}). In the formalism of \cite{MTV97}, 
an $\ell$-node (serial) episode, $\alpha$, is defined as a tuple, $(V_\alpha,<_\alpha,g_\alpha)$, where
$V_\alpha = \{v_1, \ldots, v_\ell\}$ denotes a collection of nodes, $<_\alpha$ denotes a {\em total
order}\footnote{In general, $<_\alpha$ can be any partial order over $V_\alpha$. In this paper, we only 
consider the case of episodes with a total order over $V_\alpha$. In \cite{MTV97}, these are referred
to as {\em serial} episodes.}  such that $v_{i} <_\alpha v_{i+1},\ i=1,\ldots,(\ell-1)$. If
$g_\alpha(v_j)=A_{i_j}, j=1,\ldots,\ell$, we use the graphical notation $(A_{j_1}\rA \cdots \rA
A_{j_\ell})$ to represent $\alpha$. An occurrence of $\alpha$ in event stream, $s=\langle (E_1, t_1), (E_2,t_2), 
\ldots, (E_n,t_n) \rangle$, is a map $h\::\:V_\alpha\rA\{1,\ldots,n\}$ such that
(i)~$E_{h(v_j)} = g(v_j)\ \forall v_j\in V_\alpha$, and (ii)~for all $v_i <_\alpha v_j$ in $V_\alpha$, the times
of occurrence of the $i^\mathrm{th}$ and $j^\mathrm{th}$ event in the occurrence are related
according to $t_{h(v_i)}\leq t_{h(v_j)}$ in $s$. 

\begin{example}
Consider a 3-node episode $\alpha = (V_\alpha,<_\alpha,g_\alpha)$, such that,
$V_\alpha=\{v_1,v_2,v_3\}$, $v_1 <_\alpha v_2$, $v_2<_\alpha v_3$ and $v_1<_\alpha v_3$, and
$g_\alpha(v_1)=A$, $g_\alpha(v_2)=B$ and $g_\alpha(v_3)=C$. The graphical representation for this
episode is $\alpha = (A\rA B\rA C)$, indicating that in every occurrence of $\alpha$, an event
of type $A$ must appear before an event of type $B$, and the $B$ must appear before an event of type
$C$. For example, in sequence (\ref{eq:ex-seq}), the subsequence $\langle
(A,1),(B,3),(C,9)\rangle$ \label{ex:example-episode} constitutes an occurrence of $(A\rA B\rA C)$.
For this occurrence, the corresponding $h$-map is given by, $h(v_1)=1$, $h(v_2)=2$ and $h(v_3)=5$.
\end{example}

In the episode formalism reviewed so far, the exact time-stamps on the events in the data stream
are only used to check the time-order of events
constituting an episode occurrence. There are many ways to incorporate explicit time-duration
constraints in episode occurrences (like the windows-width constraint of \cite{MTV97}, or the
dwelling time constraint of \cite{LSU07a}). Episodes with inter-event {\em gap}
constraints were introduced in \cite{PSU08}. For example, the framework of \cite{PSU08} can express
the temporal pattern ``$B$ must follow $A$ {\em within} 5 time-ticks and $C$ must follow $B$ {\em
within} 10 time-ticks.'' Such a pattern is represented using the graphical notation, $(A \stackrel{[0\sm5]}{\rrA} B \stackrel{[0\sm10]}{\rrA} C)$.
In this paper, we use a simple sub case of the inter-event gap constraints,
in the form of fixed inter-event time-delays. Here, each inter-event time constraint is represented
by a {\em single} delay rather than a range of delays. We will refer to such episodes
as {\em fixed-delay episodes}. For example, $(A\stackrel{5}{\rA} B \stackrel{10}{\rA}C)$
represents a fixed-delay episode, every occurrence of which must comprise an $A$, followed by a $B$
{\em exactly} 5 time-ticks later, which in-turn is followed by a $C$ {\em exactly} 10 time-ticks later.

\begin{definition}
An $\ell$-node {\em fixed-delay episode} is defined as a pair, $(\alpha, \scrD)$, where $\alpha=
(V_\alpha, <_\alpha, g_\alpha)$ is the usual (serial) episode of \cite{MTV97}, and $\scrD
=(\delta_1,\ldots,\delta_{\ell-1})$ is a sequence of $(\ell-1)$ non-negative delays. Every
occurrence, $h$, of the fixed-delay episode in an event sequence, $s=\langle (E_{j_1},t_1), \ldots,
(E_{j_n},t_n) \rangle$, must satisfy the inter-event constraints, $\delta_i = (t_{h(v_{i+1})} - t_{h(v_i)}),\ i=1,\ldots,(\ell-1)$.
$(A_{j_1} \stackrel{\delta_1}{\rrA} \cdots \stackrel{\delta_{\ell-1}}{\rrA} A_{j_\ell})$ is the
graphical notation for inter-event episode, $(\alpha,\scrD)$, where $A_{j_i} = g_\alpha(v_i),\ i=1,\ldots,\ell$.
\label{def:fixed-delay-episode}
\end{definition}
The framework of frequent episode discovery requires the notion of episode frequency. This can be
defined in many ways~\cite{MTV97,laxman06}. In this paper, we use the notion of {\em distinct
occurrences} to define the frequency of fixed-delay episodes.
\begin{definition}
Two occurrences, $h_1$ and $h_2$, of a fixed-delay episode, $(\alpha,\scrD)$, are said to be {\em
distinct}, if they do not share any events in the data stream, $s$. Given a user-defined, $W>0$,
{\em frequency} of $(\alpha,\scrD)$ in $s$, denoted $f_s(\alpha,\scrD,W)$,
is defined as the total number of distinct occurrences of $(\alpha,\scrD)$ in $s$ that
terminate strictly after $W$.
\label{def:distinct-occurrences}
\end{definition}
In general, counting distinct occurrences of episodes suffers from computational
inefficiencies \cite{laxman06}. (Each occurrence of an episode $(A\rA B\rA C)$ is a
substring that looks like $A\ast B\ast C$, where $\ast$ denotes a {\em variable-length} don't-care,
and hence, counting all distinct occurrences in the data stream
can require an unbounded number of automata for each episode). However, in case of fixed-delay
episodes, it is easy to track distinct occurrences efficiently. For example, when counting frequency
of $(A\stackrel{3}{\rrA}B\stackrel{5}{\rrA}C)$, if we encounter an $A$ at time $t$, to recognize an occurrence
involving this $A$ we only need to check for a $B$ at time $(t+3)$ and for a $C$ at time $(t+8)$. In
addition to being attractive from an efficiency point-of-view, we will show next in
Sec.~\ref{sec:dbn-marginals-from-episode-frequencies} that the distinct occurrences-based frequency 
count for fixed-delay episodes will allow us to interpret relative frequencies as probabilities of
DBN marginals. (Note that the $W$ in {\em Definition~\ref{def:distinct-occurrences}}
is same as length of the history window used in the constraint {\bf A1}. Skipping any occurrences
terminating in the first $W$ time-ticks makes it easy to normalize the frequency count into a
probability measure).

\subsection{Marginals from episode frequencies}
\label{sec:dbn-marginals-from-episode-frequencies}

In Sec.~\ref{sec:optimal-dbn-structure} we derived the optimal DBN
structure for an event sequence under constraints {\bf A1} and {\bf A2}, and assumption {\bf A3}.
The main result was that for $t>W$, the (optimal) parents of a node,
$X_j(t)$, corresponds to the subset, $\pi_j(t)\subset \{X_k(\tau):\:\: 1\leq k\leq
M; (t-W)\leq \tau < t\}$, which maximizes the mutual information, $I[X_j(t)\:;\:\pi_j(t)]$. In this
section, we describe how to compute this mutual information from the frequency counts of fixed-delay
episodes.

Every subset of event-indicators in the network is associated with a fixed-delay episode. 
\begin{definition}
Let $\{X_j(t)\::\:j=1,\ldots,M;\ t=1,\ldots, T\}$ denote the collection of event-indicators
used to model event stream, $s=\langle(E_1,t_1),\ldots (E_n,t_n)\rangle$, over alphabet,
$\scrE=\{A_1,$ $\ldots,A_M\}$. Consider
an $\ell$-size subset, $\scrX=\{X_{j_1}(t_1),\ldots,$ $X_{j_\ell}(t_\ell)\}$, of these indicators, and
without loss of generality, assume $t_1\leq \cdots \leq t_\ell$. Define the $(\ell-1)$ inter-event
delays in $\scrX$ as follows: $\delta_j = (t_{j+1}-t_j)$, $j=1,\ldots,(\ell-1)$. The fixed-delay episode,
$(\alpha(\scrX),\scrD(\scrX))$, that is associated with the subset, $\scrX$, of event-indicators is
defined by $\alpha(\scrX) = (A_{j_1}\rA \cdots \rA A_{j_\ell})$, and
$\scrD(\scrX)=\{\delta_1,\ldots,\delta_{\ell-1}\}$. In graphical notation, the fixed-delay
episode associated with $\scrX$ can be represented as follows:
\begin{equation}
(\alpha(\scrX),\scrD(\scrX)) = (A_{j_1} \stackrel{\delta_1}{\rA} \cdots \stackrel{\delta_{\ell-1}}{\rA} A_{j_\ell}).
\end{equation}
\label{def:rvset-episode-association}
\end{definition}
For computing mutual information using Eq.~(\ref{eq:mutual-information}), we need the
marginals of various subsets  of event-indicators in the network. Given 
a subset like $\scrX=\{X_{j_1}(t_1), \ldots, X_{j_\ell}(t_\ell)\}$, we
need estimates for probabilities of the form, $P[X_{i_1}(t_1) = a_1, \ldots,$ $X_{i_\ell}(t_\ell) = 
a_\ell]$, where $a_j\in\{0,1\}$, $j=1,\ldots,\ell$. The fixed-delay episode,
$(\alpha(\scrX),\scrD(\scrX))$, that is associated with $\scrX$ is given by 
{\em Definition~\ref{def:rvset-episode-association}} and its frequency in the data stream, $s$,
is denoted by $f_s(\alpha(\scrX),\scrD(\scrX),W)$ (as per {\em
Definition~\ref{def:distinct-occurrences}}) where $W$ denotes length of history window as per
{\bf A1}. Since an occurrence of the fixed-delay episode, $(\alpha(\scrX),\scrD(\scrX))$,
can terminate in each of the $(T-W)$ time-ticks in $s$, the probability of an 
all-ones assignment for the random variables in $\scrX$ is given by:
\begin{align}
P[X_{i_1}(t_1)=1,\ldots, X_{i_\ell}(t_\ell)=1] = \frac{f_s(\alpha(\scrX),\scrD(\scrX),W)}{T-W}
\end{align}
For all other assignments (i.e.~for assignments that are not all ones) we need to use the inclusion-exclusion
formula to obtain corresponding probabilities. The inclusion-exclusion principle has been used
before in data mining, e.g.~for approximating queries using frequent
itemsets \cite{seppanen06}. The idea is to obtain exact or approximate frequency counts for arbitrary
boolean queries using only counts of {\em frequent} itemsets in the data. In our case, counting
distinct occurrences of fixed-delay episodes facilitates use of the inclusion-exclusion formula
for obtaining the probabilities needed in the mutual information expression of
Eq.~(\ref{eq:mutual-information}).
Consider the set, $\scrX=\{X_{j_1}(t_1),\ldots,X_{j_\ell}(t_\ell)\}$, of $\ell$ event-indicators, and let
$(a_1,\ldots,a_\ell)$, $a_j\in\{0,1\},\ j=1,\ldots,\ell$, be any general assignment for
the event-indicators in $\scrX$. Let $\scrU\subset\scrX$ denote the set of indicators out of
$\scrX$ for which corresponding assignments in $\scrA$ are all 1's, i.~e.~$\scrU = 
\{X_{j_k}\in\scrX\::\:k \mathrm{\ s.t.\ } a_k=1\mathrm{\ in\ } \scrA, 1\leq k\leq \ell\}$. The inclusion-exclusion
formula can now be used to compute the probabilities as follows:
\begin{align}
P[X_{j_1}&=a_1, \ldots, X_{j_\ell}=a_\ell] \nonumber\\
&= \sum_{\begin{array}{c}\scrY\mathrm{\ s.t.\ }\\
\scrU\subseteq\scrY\subseteq\scrX\end{array}} (-1)^{|\scrY\setminus \scrU|} \left( \frac{f_s(\scrY)}{T-W} \right) 
\label{eq:inclusion-exclusion}
\end{align}
where we use $f_s(\scrY)$ as short-hand for $f_s(\alpha(\scrY),\scrD(\scrY),W)$, the distinct
occurrences-based frequency (cf.~{\em Definition~\ref{def:distinct-occurrences}}) of the fixed-delay
episode, $(\alpha(\scrY),\scrD(\scrY))$. It is possible to verify that summing the expression in
Eq.~(\ref{eq:inclusion-exclusion}) over all possible binary $\ell$-tuples, $(a_1,\ldots,a_\ell)$,
always yields 1. Thus, in case of fixed-delay episodes, suitably-normalized
frequency counts can be regarded as corresponding marginal probabilities.

%
%
%
%

\section{Excitatory networks}
\label{sec:excite}
%

In this section, we describe a restricted class of networks, called
{\em excitatory networks} where only certain kinds of conditional dependencies among nodes are
perimitted. In general, each event-type from the
alphabet can have some propensity of random occurrence (and this of course, can be different for
different event-types). In so-called excitatory networks the only way to increase 
the propensity of occurrence of an event-type is by the occurrence of specific event-types 
at specific delays in the immediate past. For example, we can have a conditional dependency
such as, ``whenever $A$, $B$ and $C$ occur (say) 2 time-ticks apart, the probability of
occurrence of $D$ increases.'' No other kinds of conditional dependencies are permitted in
excitatory networks. In other words, one or more event-types
cannot {\em increase} the propensity of another by {\em not} occurring in the data stream. This kind of
constraint appears naturally in neuroscience, where one is interested in unearthing conditional
dependency relationships among neuron spiking patterns. There are several regions in the
brain that are known to exhibit predominantly excitatory characteristics and in these cases a neuron
cannot increase the firing rate of another by not firing.

In the context of DBN structure learning, this restriction translates to event-types frequently
appearing alongside their respective parents (with specific delays). This motivates the use of {\em frequent}
episodes to restrict the search space for parent sets. We note that, in general, it is difficult
to infer an optimal network by inspection of the set of frequent episodes and/or the
association rules that can be derived from it. This is because, there can be several frequent
episodes/rules ending in the same event-type and these may even be of different sizes. There is no obvious
way to systematically analyze frequencies and confidence values among these candidates to pick
the optimal parent(s) in the network. In Sec.~\ref{sec:results}, we will present some examples to illustrate the shortcomings of our frequency based approach in detecting higher order causative relationships.

\section{Algorithms}


In Secs.~\ref{sec:optimal-dbn-structure}-\ref{sec:episodes}, we developed the formalism for learning
an optimal DBN structure from event streams by using distinct occurrences-based counts of fixed-delay
episodes to compute the DBN marginal probabilities. The top-level algorithm for discovering the
network is to fix any time $t>W$ and consider each $X_i(t)$, $i=1,\ldots,M$, in-turn. The best
parent set, $\pi_i(t)$, for $X_i(t)$ is selected by searching over subsets of event-indicators
(in the $W$-length history of $X_i(t)$) for the one that maximizes the mutual information,
$I[X_i(t)\:;\:\pi_i(t)]$. To compute this mutual information we use the formalism of fixed-delay
episodes. The marginal probabilities required in the computation of $I[X_i(t)\:;\:\pi_i(t)]$ are
obtained using Eq.~(\ref{eq:inclusion-exclusion}) which basically uses the frequencies of
appropriate fixed-delay episodes in an inclusion-exclusion formula. Finally, since we are looking
for only excitatory connections, we restrict our search space to {\em frequent} fixed-delay
episodes. Thus the first step in our DBN learning algorithm is to detect all frequent fixed-delay
episodes with a span less than $W$. This is described in Sec.~\ref{sec:frequent-episode-discovery}.
These frequent episodes (along with their respective
frequencies) are input to the actual parents-search algorithm (which constructs different parents
and picks the one with highest mutual information).
This is explained next in Sec.~\ref{sec:learning-dbn-structure}


\begin{algorithm}[!ht]
\floatname{algorithm}{Procedure}
\caption{$pattern\_grow(\alpha,\scrD_{\alpha},\scrS)$}
\label{alg:pattern-grow}
\begin{algorithmic}[1]
\REQUIRE $N$-node episode $\alpha=\langle E_{j_1}\stackrel{\delta_1}{\longrightarrow}\ldots E_{(N)}\rangle$ and event sequence $\scrS=\{(E_i, t_i)\}, i \in \{1\ldots{n}\}$, Length of history window $W$.
\STATE $\Delta = W - span(\alpha)$
\FORALL{$A \in \scrE$}
	\FOR{$\delta = 0 \mbox{ to }\Delta$}
		\IF{$\delta = 0$ \textbf{and} $\alpha[1]>A$}
			\STATE \textbf{continue}
		\ENDIF
		\STATE $\beta = A \stackrel{\delta}{\rA} \alpha$
		\STATE \COMMENT{Obtain count of $\beta$ in projected data sequence $\scrD_{\alpha}$}
		\FORALL{$i \in \scrD_{\alpha}$}
			\STATE $(E_i, t_i) = \scrS[i]$
			\IF{$\exists j$ such that $E_j = A$ and $t_i - t_j = \delta$}
				\STATE Increment $\beta.count$
				\STATE $\scrD_{\beta} = \scrD_{\beta} \cup \{j\}$
			\ENDIF
		\ENDFOR
		\IF{$\beta.count > \theta$}
			\STATE Add $\beta$ to $\scrF(k+1)$
			\IF{$|\beta| < k+1$}
				\STATE $pattern\_grow(\beta,\scrD_{\beta},\scrS)$
			\ENDIF
		\ENDIF
	\ENDFOR
\ENDFOR
\end{algorithmic}
\end{algorithm}

\subsection{Discovering fixed-delay episodes}
\label{sec:frequent-episode-discovery}

However for long patterns and low support thresholds, an Apriori like algorithm incurs substantial costs and further must repeatedly scan the entire database for each level. 
In Algorithm~\ref{alg:pattern-grow} we present
a pattern-growth based algorithm for mining frequent episodes (with fixed delays).

\begin{figure}[!htbp]
	\centering
		\includegraphics[width=0.6\columnwidth]{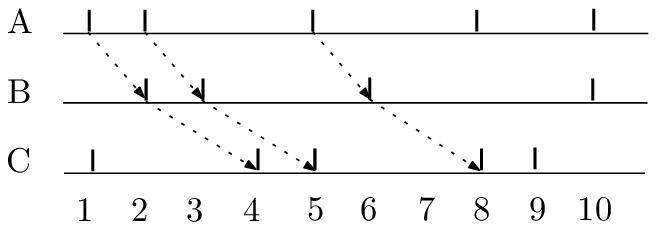}
		\caption{An event sequence showing 3 distinct occurrences of the episode $A\stackrel{1}{\rightarrow}B\stackrel{2}{\rightarrow}C$.}
	\label{fig:seq}
\end{figure}
The pattern-growth procedure given in Algorithm~\ref{alg:pattern-grow} takes as input an episode $\alpha$, a set of integers $\scrD_{\alpha}$, and the actual event sequence $\scrS$. $\scrD_{\alpha}$ is a set time stamps $t_i$ such that there is an occurrence of $\alpha$ starting at $t_i$ in $\scrS$. For example, say at level 1 we have $\alpha=(C)$, then $\scrD_{(C)}$ = $\{1,4,5,8,9\}$ for the event sequence $\scrS$ shown in Fig~\ref{fig:seq}. The algorithm proceeds by obtaining counts for all episodes $\beta$ generated by extending $\alpha$ e.g. $B\stackrel{0}{\rA}C$, $\ldots$, $A\stackrel{5}{\rA}C$ etc. For an episode say $\beta = B\stackrel{2}{\rA}C$, the count is obtained looking for an occurrence of event $B$ at time $t_j = t_i-2$ where $t_i\in\scrD_{B}$. In the example $t_j$=$\{2,3,6\}$. The number of such occurrences gives the count of $B\stackrel{2}{\rA}C$. At every step the algorithm tries to grow an episode with count $>\theta$ otherwise stops.


\subsection{Learning DBN structure}
\label{sec:learning-dbn-structure}



In Sec.~\ref{sec:optimal-dbn-structure} we derived the conditions of an optimal DBN structure. Based on our assumptions, it suffices to look at a one time instance $t>W$ and ascertain the parents of each of the M nodes $\XB(t)$. Translation invariance then automatically allows us to fix the parents of all other $\XB(t)$'s at time instance $(t>W)$.

The algorithm considers one node corresponding to each event-type in the alphabet in-turn, and searches for its best parent set. Based on our discussion in Sec.~\ref{sec:excite}, we restrict the search space to only frequent episodes (ending in the event-type). If there are no frequent episodes ending in an event-type, we declare it as a root node. In addition from mutual information theory we know that in general
\begin{equation}
I[X; \scrY] \geq I[X; \scrZ], \forall \scrZ \subset \scrY
\label{eq:mi-rel}
\end{equation}
where $X$ is a single random variable and $\scrY$ and $\scrZ$ are sets of random variables. But on removing the nodes that do not contribute in causing $X_j$ from the parent set $\pi_j$, the mutual information decreases only slightly. This is the basis of Algorithm~\ref{alg:dbn-learn} where at each step the parent set of size $m$, $\pi^m_j(t)$ is chosen from the frequent episode of size $(m+1)$ which gives the highest mutual information with $X_j(t)$. In the case where a parent set of size $n$, $\pi^n_j(t)$ with $n>m$ has already been selected for $X_j(t)$ in an earlier iteration, the set $\pi^m_j(t)$ replaces $\pi^n_j(t)$ if $I[X_j(t), \pi^m_j(t)] - I[X_j(t), \pi^m_j(t)] < \epsilon$ and $\pi^m_j(t) \subset \pi^n_j(t)$.

In addition $\pi^m_j$ replaces $\pi^n_j$ if $I[X_j, \pi^m_j] > I[X_j, \pi^m_j]$ for $\pi^m_j \not\subset \pi^n_j$. Therefore using the $\epsilon$ criteria we can iteratively remove nodes from the parent set that do not contribute in the information theoretic sense towards the cause of $X_j$.

\begin{algorithm}[ht]
\begin{algorithmic}[1]
\STATE /* Initialize */
\STATE $h = \{\}$ /* Empty hash-map */
\FOR{$i = k \mbox{ down to } 1$}
	\FORALL{$\alpha \in F_{i+1} $}
		\STATE /* $F_{i+1}$: frequent episodes of size $i+1$ */
		\STATE {$A=$ Last event of $\alpha$}
		\STATE $par = prefix(\alpha)$
		\STATE /* $par$: first $|\alpha-1|$ nodes chosen as candidate parents of A with delays corresponding to inter-event gaps in $\alpha$ */
		\IF{$A \notin h$}
			\STATE $h(A) = (par, MI(A, par), i)$
		\ELSE
			\STATE $(par_{prev}, mi_{prev}, level) = h(A)$
			\IF{$level = i + 1$}
				\STATE $mi = MI(A, par)$
				\IF{$|mi-mi_{prev}| < \epsilon$ \textbf{or} $mi > mi_{prev}$ } 
					\STATE $h(A)=(par, mi, i)$
				\ENDIF
			\ELSIF{$level = i$}
				\STATE $mi = MI(A, par)$
				\IF{$mi > mi_{prev}$}
					\STATE $h(A)=(par, mi, i)$
				\ENDIF
			\ENDIF
		\ENDIF
	\ENDFOR
\ENDFOR
\STATE \textbf{Output}: DBN = $\{(A, h[A].par), \forall A \}$ gives the DBN 
\end{algorithmic}
\caption{DBN learning from frequent episodes}
\label{alg:dbn-learn}
\end{algorithm}

The time complexity of computing mutual information with a parent set of size $k$ is $O(2^k)$ as we have to compute $2^k$ value assignments to all the nodes in the parent set (which take values 0 or 1). But since $k$ is a user-supplied
parameter (and assumed constant for a given run of Algorithm~\ref{alg:dbn-learn}), the time complexity is output-sensitive
with linear dependence on the number of frequent episodes $O(|F_{i+1}|)$at each level $i$.


\section{Experimental Results}
We present results on data gathered from both mathematical models of spiking neurons
as well as real neuroscience datasets.
\label{sec:results}

\subsection{Data generation model}
Our approach here is to model each neuron as an inhomogeneous Poisson process\footnote{simulator courtesy Mr. Raajay,
M.S. Student, IISc, Bangalore.} whose firing rate is a 
complex function of the input received by the neuron:
\begin{equation}
\lambda_{i}(t) = \frac{\lambda}{1 + exp(-I_i(t) + \theta)}
\label{eq:simulator}
\end{equation}
Eq~\ref{eq:simulator} gives the firing rate of the $i^{th}$ neuron at time $t$. The network inter-connect allowed by this model gives it the amount of sophistication required for simulating higher-order interactions. More importantly, the model allows for variable delays which mimic the delays in conduction pathways of real neurons. 
\begin{equation}
I_i(t) = \sum_{j}{\beta_{ij}Y_{j(t-\tau_{ij})}} + \ldots + \sum_{ij\ldots{l}}{\beta_{ij\ldots{l}}Y_{j(t-\tau_{ij})}\ldots Y_{l(t-\tau_{il})}}
\label{eq:higer-order}
\end{equation}
In Eq~\ref{eq:higer-order}, $Y_{j(t-\tau_{ij})}$ is the indicator of the event of a spike on $j^{th}$ neuron $\tau_{ij}$ time earlier. The higher order terms in the input contribute to the firing rate only when the $i^{th}$ neuron received inputs from all the neurons in the term  with corresponding delays. With suitable choice of parameters $\beta_{(.)}$ one can simulate a wide range of networks. 

\subsection{Types of Networks}
In this section we demonstrate the effectiveness of our approach in unearthing different types of networks. Each of these networks was simulated by setting up the appropriate inter-connections, of suitable order, in 
our mathematical model.

\textit{Causative Chains}: These are
simple first order interactions forming linear chains. The parent set
for each random variable is a single variable. Observe that this class
includes loops in the underlying graph that 
would be `illegal' in a static Bayesian network formulation.
A causative
chain is perhaps the easiest scenario for DBN inference. 
Here a network with 50 nodes is simulated for 60 sec on the 
multi-neuronal simulator, where the
conditional probability is varied form 0.4 to 0.8.
For a reasonably high conditional probability (0.8), we obtain 100\%
precision for a reasonablly wide
range  of the frequency threshold ($[0.002,0.038]$).
The recall is also similarly high but drops a bit toward higher values of
the frequency threshold. For the low conditional probability scenario,
the number of frequent episodes mined drops to zero and hence no network is
found (implying both a precision and recall of 0).
A similar experiment was conducted for different values of parameter 
$\epsilon$ and $k$. For this particular network, the results are
robust as there are only first order interactions.

\begin{figure}[!htbp]
	\centering
		\subfigure[scriptsize][Causative chains]
		{\label{fig:net-1}\includegraphics[height=0.45\columnwidth]{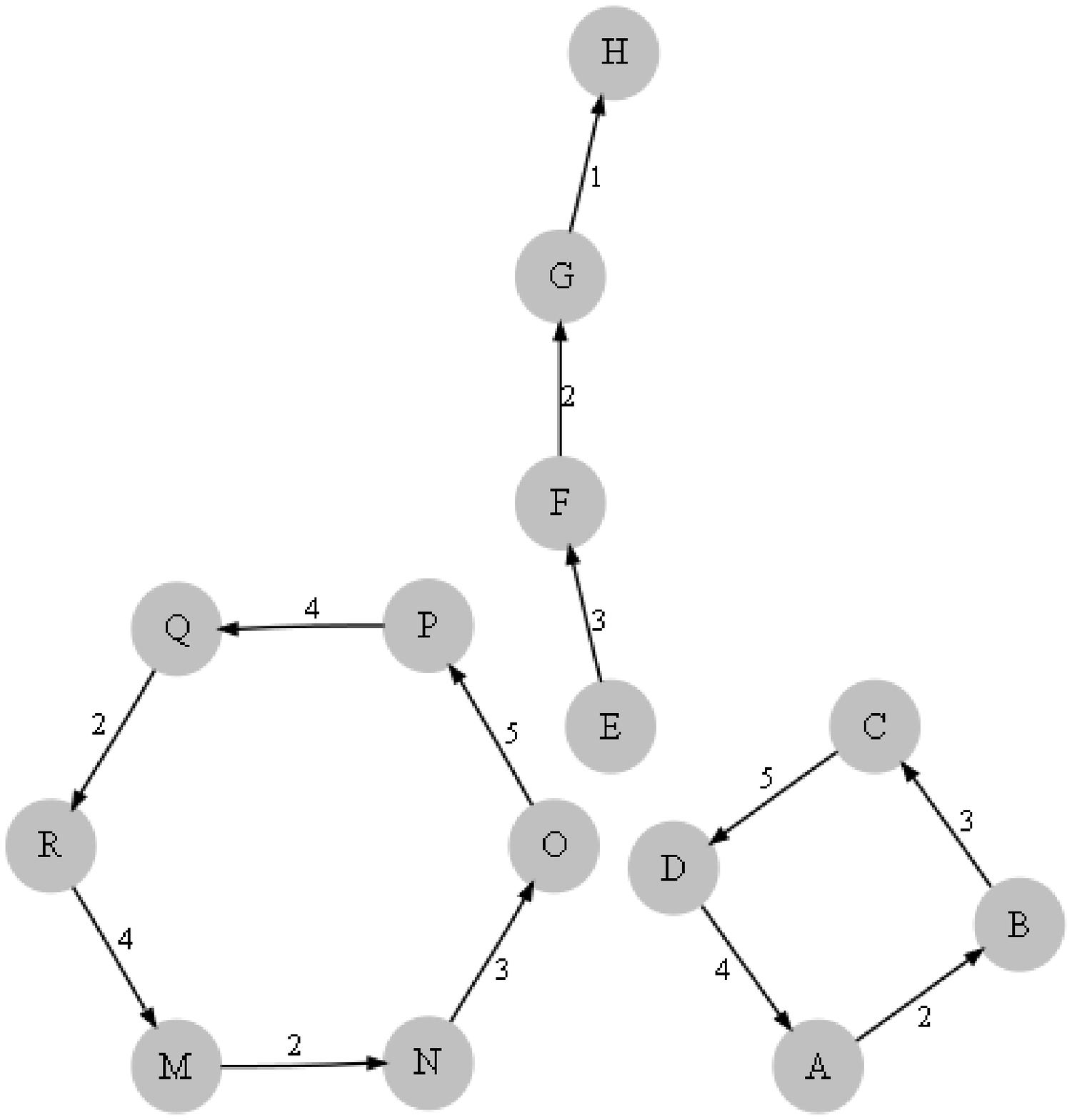}}
		\subfigure[scriptsize][Higher-order causative chains]
		{\label{fig:net-2}\includegraphics[height=0.45\columnwidth]{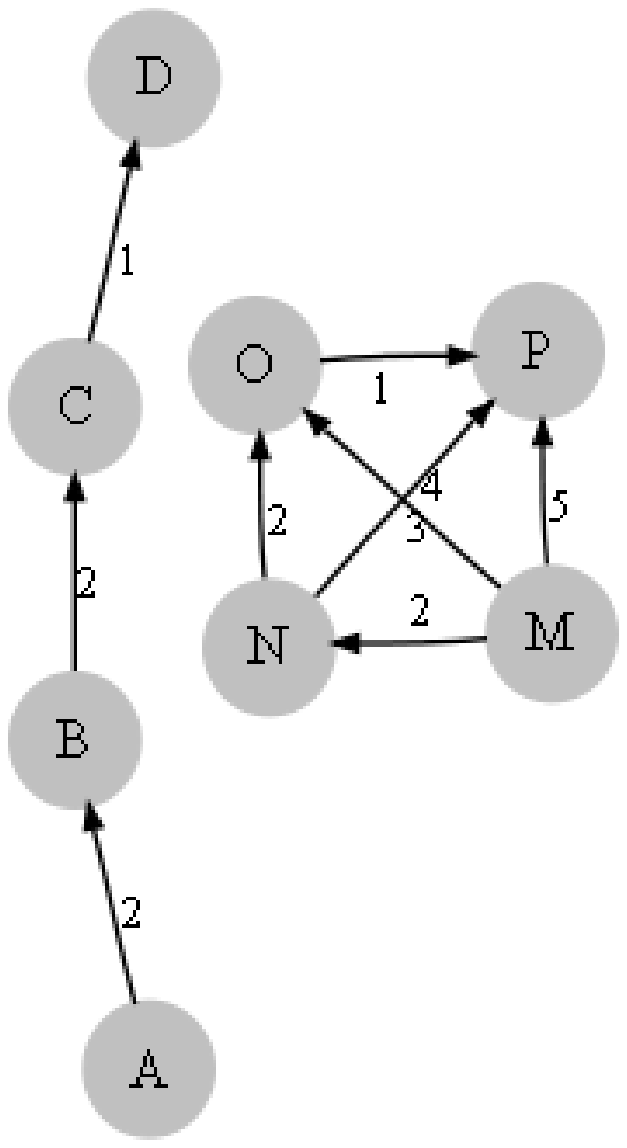}}
		\subfigure[scriptsize][Syn-fire Chains]
		{\label{fig:net-3}\includegraphics[height=0.55\columnwidth]{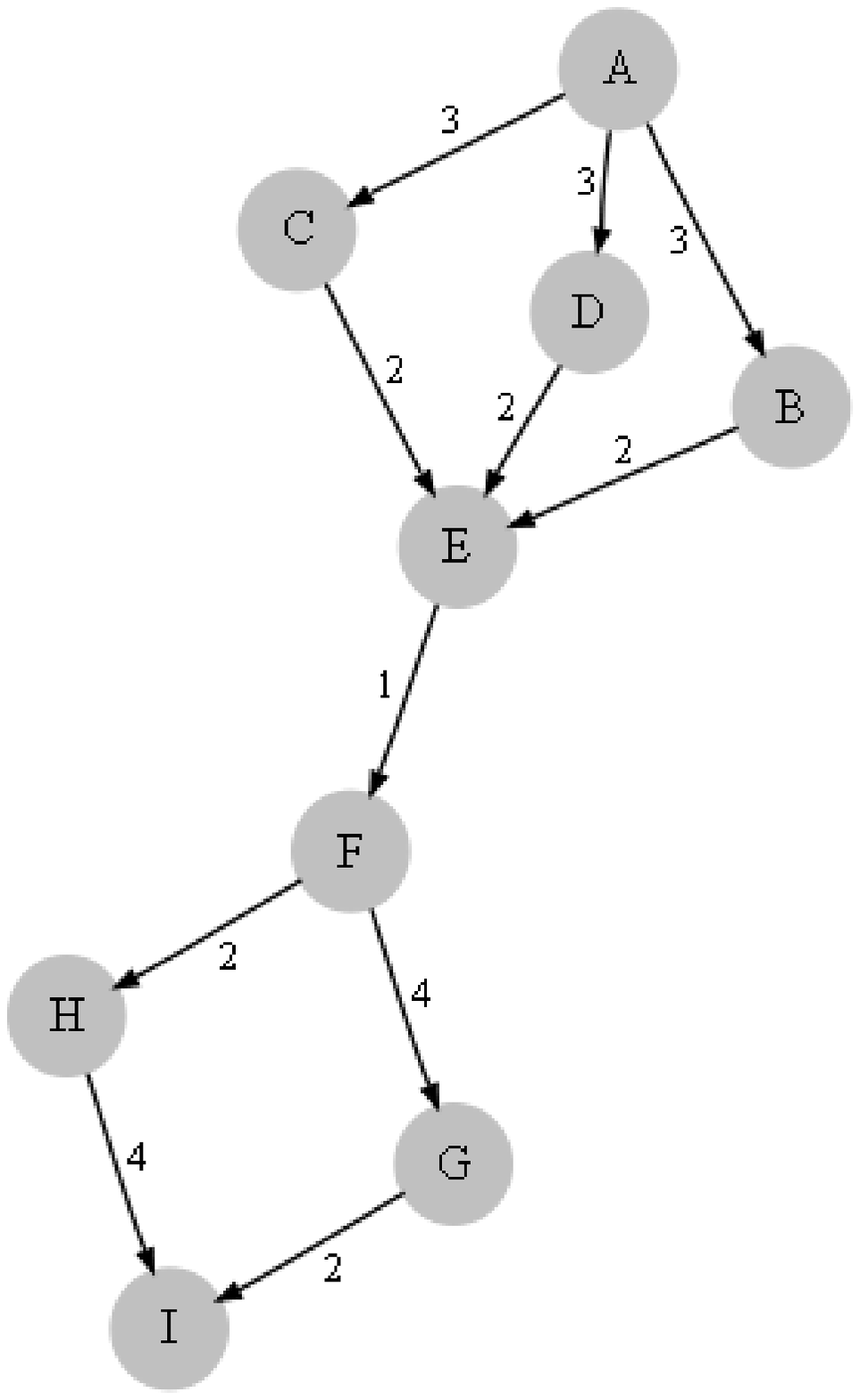}}
		\subfigure[scriptsize][Polychronous Circuits]
		{\label{fig:net-4}\includegraphics[height=0.35\columnwidth]{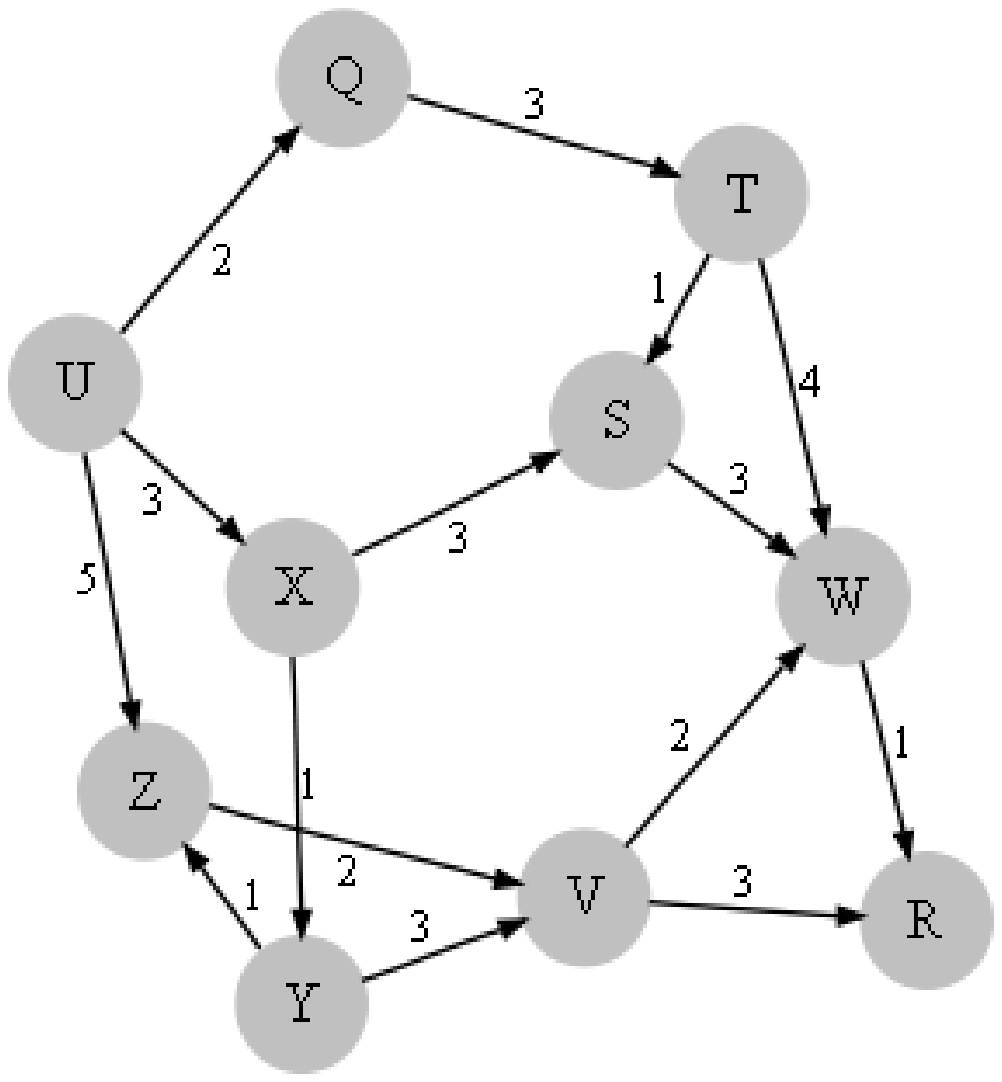}}
	\caption{Four classes of DBNs investigated in our experiments.}
\end{figure}

\textit{Higher-order causative chains}: A higher-order chain is one 
where parent sets are not restricted to be of cardinality one.
In the example network of Fig.~\ref{fig:net-2} we have two
disconnected components:
one first order causative chain formed by nodes A,B,C and D and a 
higher-order causative chain formed by M,N,O and P. In the higher order chain, 
O fires when {\it both} M and N fire with appropriate timing and 
P fires when all of M, N, and O fire. Looking only at frequent 
episodes, both $A\rA B \rA C\rA D$ and $M\rA N \rA O\rA P$ turn 
out to be frequent. But using the $\epsilon$  criteria for our algorithm 
we can distinguish the two components of the circuits to be of different
orders. Table~\ref{tab:net-2-eps} demonstrates results on this network
as $\epsilon$ is varied from $0.00001$ to $0.01$, and for the same range
of conditional probabilities as before.
A low $\epsilon$ results in a decrease of precision, e.g.,
our approach finds A and B to be parents of C. Conversely,
for higher values of $\epsilon$ 
our algorithm might reject the set M, N, P as parents of P and retain 
some subset of them.
A final observation is in reference to the times from 
Table~\ref{tab:net-2-eps}; the values presented here includes the time to
compute the mutual information terms plus the time to mine frequent patterns,
and the significant component is the latter.

\begin{table}[!htbp]
\caption{DBN results for network shown in Fig.~\ref{fig:net-2} for varying conditional probability (used in generation) and $\epsilon$ (used in mining);
base firing rate = 20Hz.}
\label{tab:net-2-eps}
\centering
\begin{tabular}{|rrrrr|}
\hline
 Cond. &    Epsilon & Time &     Recall &  Prec- \\
 prob &             & (total) &         & ision\\
 \hline
 0.8 &    0.00001 &      18.31 &        100 &         75 \\
 0.8 &     0.0001 &      18.31 &        100 &        100 \\
 0.8 &      0.001 &       18.3 &      88.89 &        100 \\
 0.8 &       0.01 &      18.31 &      77.78 &        100 \\
 0.4 &    0.00001 &      15.02 &        100 &      81.82 \\
 0.4 &     0.0001 &      14.98 &        100 &        100 \\
 0.4 &      0.001 &      14.98 &      88.89 &        100 \\
 0.4 &       0.01 &      14.98 &      66.67 &        100 \\
\hline
\end{tabular}  
\end{table}

\textit{Syn-fire Chains}: Another important pattern often seen in neuronal spike train data is that of synfire chains. This consists of groups of synchronously firing neurons strung together repeating over time. In an earlier work~\cite{PSU08}, it was noted that discovering such patterns required a combination of serial and parallel episode mining. But the DBN approach applies more naturally to mining such network structures. 

%
\textit{Polychronous Circuits}: Groups of neurons that fire in a time-locked manner with respect to each other are refer to as polychronous groups. This 
notion was introduced in~\cite{Izh06} and gives rise to an important class of patterns. Once again, our DBN formulation is a natural fit for 
discovering such groups from spike train data. An example of a polychronous circuit is show in Fig~\ref{fig:net-4} and its corresponding results in Table~\ref{tab:net-4}.

\begin{table}[!htbp]
\caption{DBN results for network shown in Fig.~\ref{fig:net-4} for different Freq. thresh. and epsilon}
\label{tab:net-4}
\centering
\begin{tabular}{|rrrrrr|}
\hline
 Cond.&      Freq. &    Epsilon &       Time &     Recall &  Prec- \\
 prob. &     Thresh &            &    (total) &            &  ision \\
\hline
   0.8 &      0.002 &     0.0005 &       22.3 &        100 &        100 \\
   0.8 &      0.014 &     0.0005 &       18.8 &         40 &        100 \\
   0.8 &      0.002 &    0.00001 &      22.81 &        100 &      53.57 \\
   0.8 &      0.002 &       0.01 &      22.91 &      93.33 &        100 \\
   0.4 &      0.002 &     0.0005 &      15.48 &      53.33 &        100 \\
   0.4 &      0.014 &     0.0005 &      15.11 &      13.33 &        100 \\
   0.4 &      0.002 &    0.00001 &      15.28 &      53.33 &      61.54 \\
   0.4 &      0.002 &       0.01 &       15.3 &      46.67 &        100 \\
\hline
\end{tabular}  
\end{table}

\begin{figure}[!htbp]
	\centering
		\includegraphics[width=0.65\columnwidth]{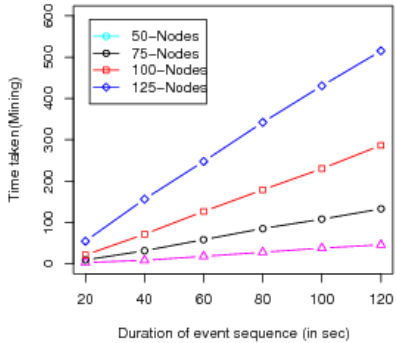}
	\caption{Plot of time taken for mining frequent episodes vs. data length in sec}
	\label{fig:scalability-mine}
\end{figure}

\begin{figure}[!htbp]
	\centering
		\includegraphics[width=0.65\columnwidth]{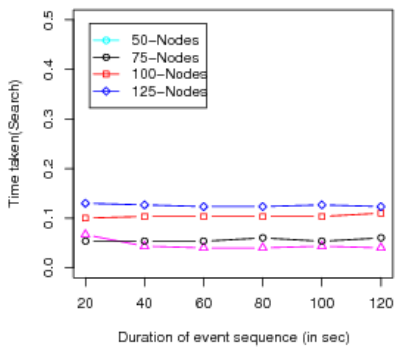}
	\caption{Plot of time taken for mutual information computation and searching for the parent set vs. data length in sec}
	\label{fig:scalability-search}
\end{figure}

\subsection{Scalability}

The scalability of our approach with respect to data length and number of variables is shown in Fig~\ref{fig:scalability-mine} and Fig~\ref{fig:scalability-search}. Here four different networks with 50, 75, 100 and 125 variables respectively were simulated for time durations ranging from 20 sec to 120 sec. The base firing rate of all the networks was fixed at 20 Hz. In each network 40\% of the nodes were chosen to have upto 3 three parents. The parameters of the DBN mining algorithm were choosen such that recall and precision are both high ($> 80\%$). It can be seen in the figures that for a network with 125 variables, the total run-time is of the order of few minutes along with recall $>$ 80\% and precision at almost 100\%.

\begin{figure}[!htbp]
	\centering
		\includegraphics[width=0.65\columnwidth]{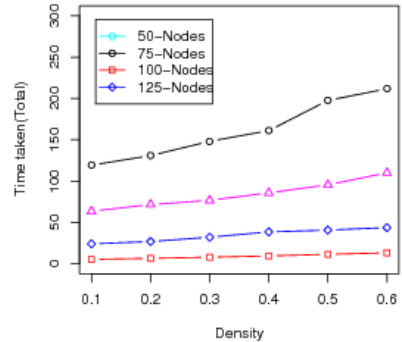}
	\caption{Plot of total time taken for DBN discovery vs. network density}
	\label{fig:scalability-density}
\end{figure}

Another way to study scalability is w.r.t. the density
of the network, defined as
the ratio of the number of nodes that are descendants for some other 
node to the total number of nodes in the network.
Fig~\ref{fig:scalability-density} shows the time taken for mining DBN when
the density is  varied from 0.1 to 0.6. 

\subsection{Sensitivity}
Finally, we discuss the sensitivity of the DBN mining algorithm to
the parameters $(\theta, \epsilon)$. To obtain 
precision-recall curves for our algorithm applied to data sequences 
with different characteristics, we vary the two parameters 
$\theta$ and $\epsilon$ in the ranges \{0.002, 0.008, 0.014, 0.026, 0.038\} 
and \{0.00001, 0.0001, 0.001, 0.01\} respectively. The data sequence for this experiment is generated from the multi-neuronal simulator using different settings of base firing rate, conditional probability, number of nodes in the 
network, and the density of the network as defined earlier. 



\begin{figure*}[!htbp]
	\centering
		\includegraphics[width=0.8\textwidth]{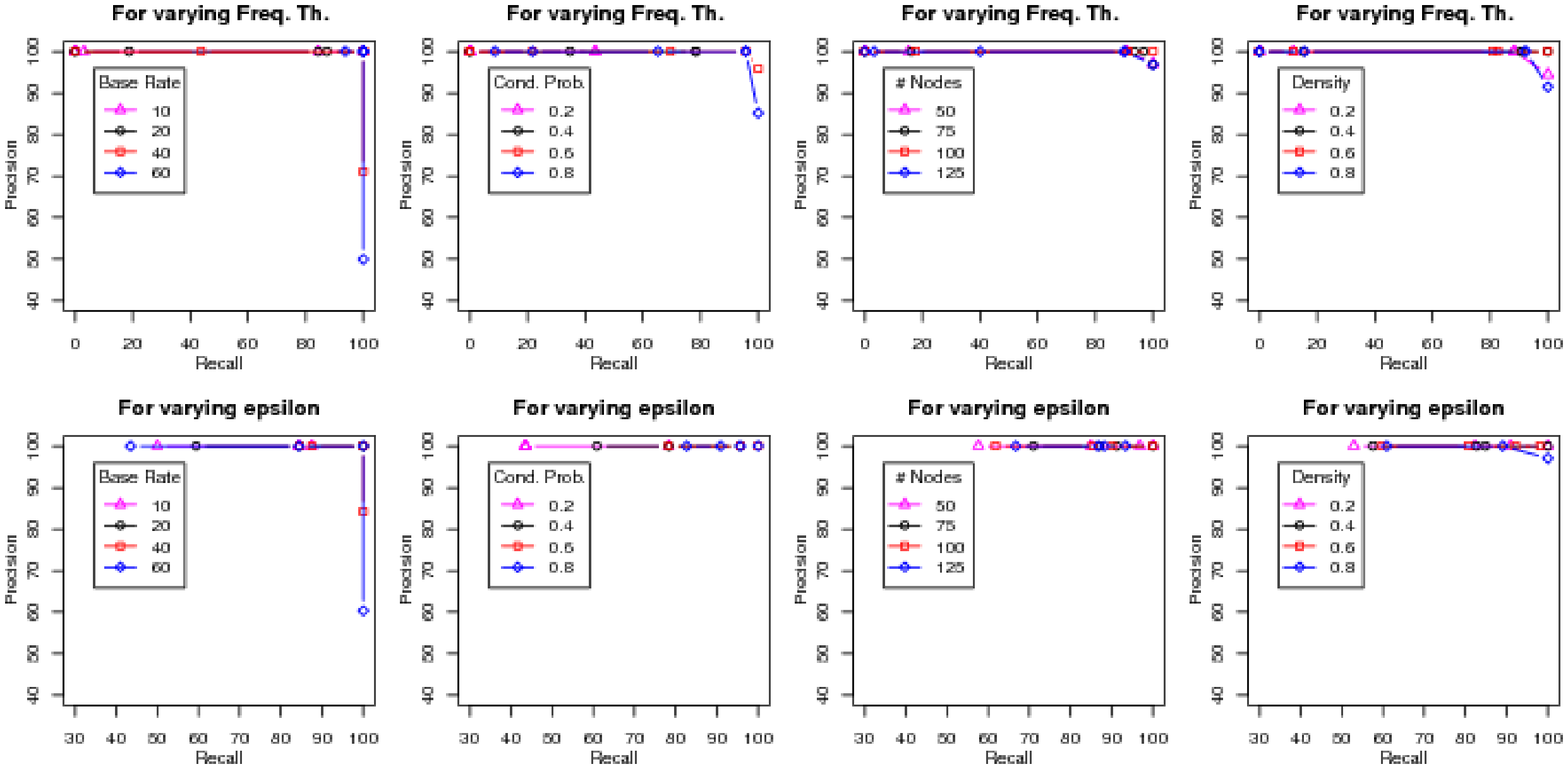}
	\caption{Precision-recall curves for different parameter values in the DBN mining algorithm.}
	\label{fig:sensitivity}
\end{figure*}

The set of precision-recall curves are shown in Fig~\ref{fig:sensitivity}. It can be seen that the proposed algorithm is effective for a wide range of parameter settings and also on data with sufficiently varying characteristics.

\subsection{Mining DBNs from MEA recordings}
\begin{figure}[!htbp]
	\centering
		\includegraphics[width=0.8in]{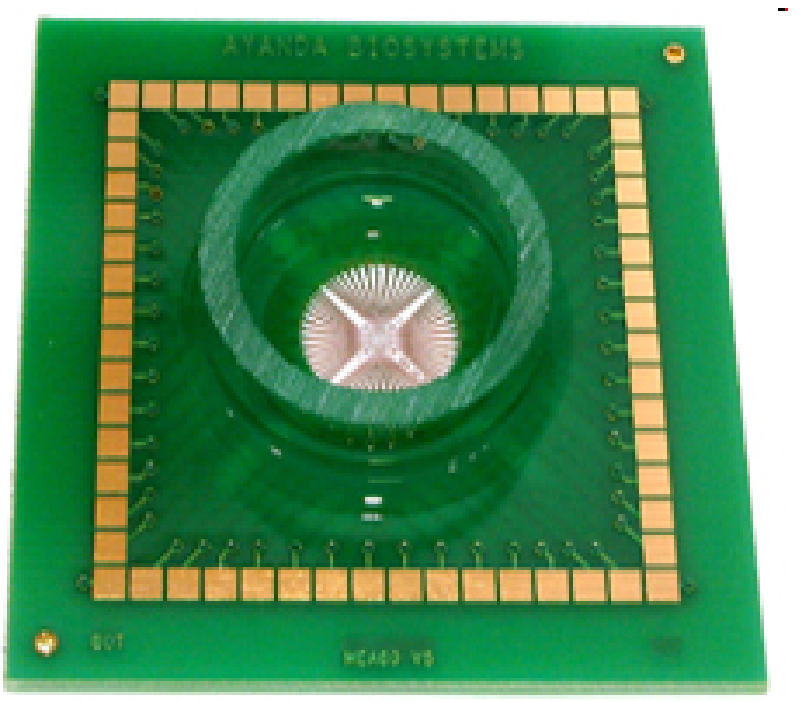}
	\caption{Micro electrode array (MEA) used to record 
spiking activity of neurons in tissue cultures.}
	\label{fig:mea}
\end{figure}

\begin{figure}[!htbp]
	\centering
		\includegraphics[width=0.8\columnwidth]{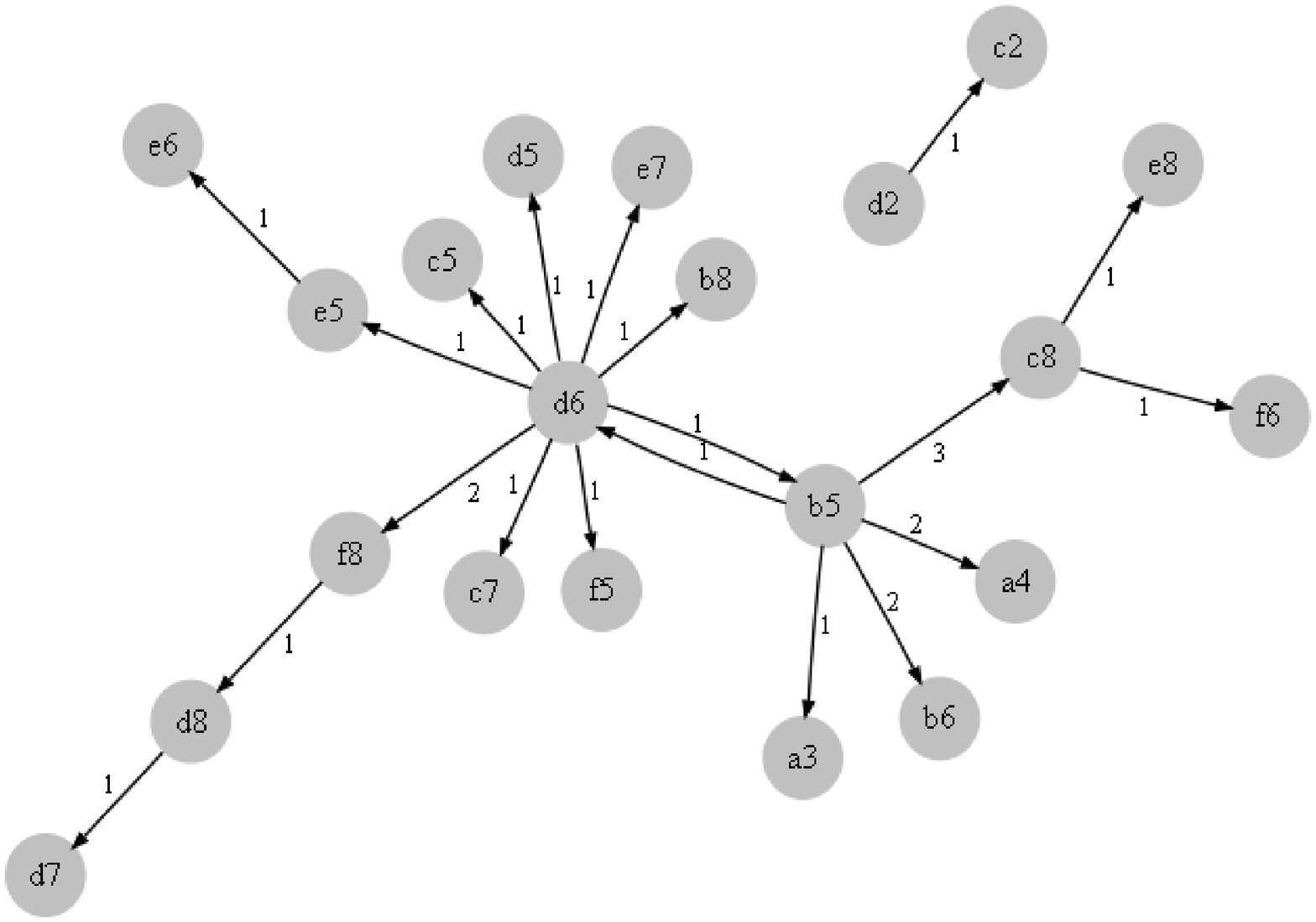}
	\caption{DBN structure discovered from neuronal spike train data.}
	\label{fig:real-dbn-1}
\end{figure}

Multi-electrode arrays (see Fig.~\ref{fig:mea}) are high throughput
ways to record the spiking activity in neuronal tissue and are hence
rich sources of event data where events correspond to specific neurons
(or clumps of neurons) being activated. We use data from
dissociated cortical cultures gathered by Steve Potter's laboratory
at Georgia Tech~\cite{Potter2006} which gathered data over several days.
The DBN shown in Fig.~\ref{fig:real-dbn-1} depicts a circuit discovered
from the first 15 min of recording on day 35 of culture 2-1. The overall mining process takes about 10 min with threshold $\theta = 0.0015$ with DBN search parameter $\epsilon = 0.0005$. 

In order to establish that this network is in fact significant we run our algorithm on several surrogate spike trains generated by replacing the neuron labels of each spikes in the real data with a randomly chosen neuron label. These surrogates are expected to break the temporal correlations in the data and yet preserve the overall summary statistics. No network structure was found in 25 such surrogate sequences. We are currently in the processes of characterizing and interpreting the usefulness of such networks found in real data.

\section{Discussion}
We have presented the beginnings of research to relate inference of DBNs 
with frequent episode mining. The key contribution here is to show how,
under certain assumptions on network structure,  data and
distributional characteristics, we are able to infer the structure 
of DBNs using the results from frequent episode mining. While our experimental
results provide convincing evidence of the efficacy of our methods, in future
work we aim to provide strong theoretical results supporting our experiences.

An open question of interest is to characterize (other)
useful classes of DBNs that have both practical relevance (like excitatory circuits) and 
which also can be tractably inferred using sufficient statistics of the form studied here.

\section{Repeatability}
Supplementary material, algorithm implementations, and results for this paper are hosted 
at http://neural-code.cs.vt.\hskip0ex edu/\hskip0ex dbn.

\end{document}